\documentclass[12pt,twoside]{article}
\usepackage[a4paper,width=150mm,top=30mm,bottom=30mm,bindingoffset=10mm]{geometry}
    \makeatletter
        \renewcommand\@makefntext[1]{%
            \ifnumgreater{\value{footnote}}{9}{%
                \noindent
                \mbox{$^{\textrm{\tiny\@thefnmark~}}$\hspace*{2.3em}}{#1}%
            }{%
                \mbox{$^{\textrm{\tiny\@thefnmark~}}$\hspace*{2.6em}}{#1}
                }%
            }%
    \makeatother %to set the correct footnote indentation
\usepackage[T1]{fontenc}
\usepackage{xcolor}
\usepackage{listings}

\lstdefinestyle{promptstyle}{
  basicstyle=\ttfamily\small,
  breaklines=true,
  keepspaces=true,
  columns=fullflexible,
  frame=single,
  framerule=0.3pt,
  rulecolor=\color{black!25},
  backgroundcolor=\color{gray!5},
  aboveskip=1.2em,
  belowskip=1.2em,
  showstringspaces=false
}

\usepackage{apacite}
\usepackage{listings}

\usepackage{booktabs}
\usepackage{array}
\usepackage{graphicx}  % for \resizebox

\usepackage{float}
\usepackage{algorithm}
\usepackage{algpseudocode}
\usepackage{amsmath}
\usepackage{amssymb}

\usepackage[utf8]{inputenc} %Standard diacritics in Romance languages (accents, umlauts)
\usepackage{times} %Uses Times New Roman font
\usepackage{wasysym}
\usepackage{ragged2e}
\usepackage{diagbox}
\usepackage[labelfont=bf, labelsep=period, font=small, justification=RaggedRight, format = plain, margin = 0pt]{caption}

\usepackage{threeparttable}
\usepackage{subcaption}
\usepackage{graphicx, xcolor}
\graphicspath{ {./images/} }
\usepackage{natbib}
\setcitestyle{notesep={: }}
\usepackage{tikz, hyperref}

\lstdefinelanguage{json}{
    basicstyle=\ttfamily\small,
    showstringspaces=false,
    breaklines=true,
    frame=single,
    string=[s]{"}{"},
    stringstyle=\color{blue},
    comment=[l]{:},
    commentstyle=\color{black},
}

\newcommand{\orcidlogo}[1]{\raisebox{-.025cm}{\includegraphics[width=10pt]{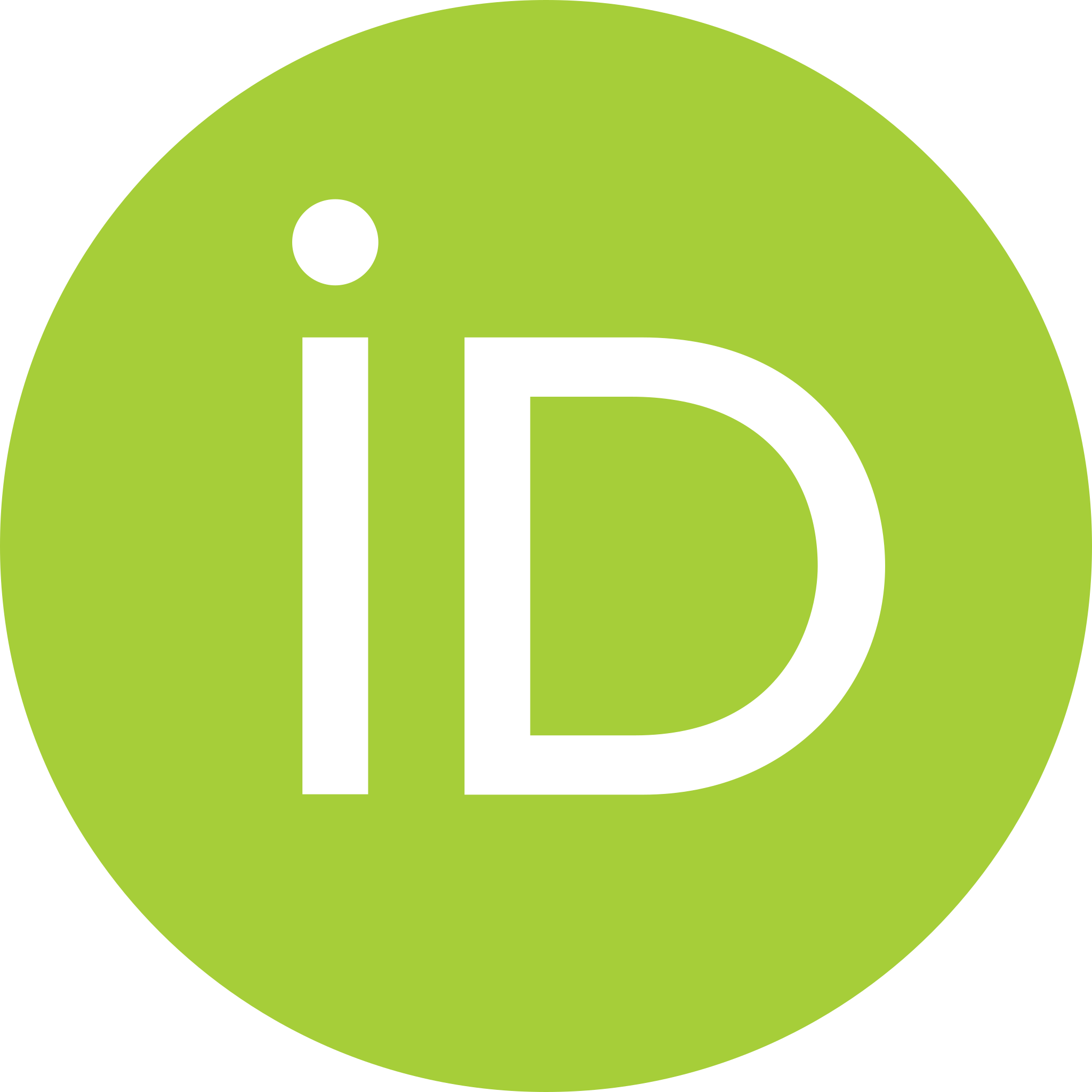}}\hspace{.5em}}
\usepackage[tiny]{titlesec}
\newcommand{\periodafter}[1]{#1.}
\titleformat{\section}{\bfseries}{\thesection\periodafter}{0.3em}{\bfseries}
\titleformat{\subsection}{\bfseries}{\thesubsection\periodafter}{0.3em}{\bfseries}
\titleformat{\subsubsection}{\itshape}{\thesubsubsection\periodafter}{0.3em}{\itshape}
\titlelabel{\thetitle.\quad}
\titlespacing*{\section}{0pt}{2em}{1em}
\titlespacing*{\subsection}{0pt}{1em}{1em}
\titlespacing*{\subsubsection}{0pt}{1em}{0pt}
\usepackage{fancyhdr}
\fancypagestyle{first}{
	\fancyhead{}
	\fancyfoot{}
}
\usepackage{tipa} %for IPA
\usepackage{phonrule} %for phonological rules
\usepackage[nocenter]{qtree} %for trees
\usepackage{expex} %for examples
\lingset{interpartskip=0mm,aboveexskip=1em,belowexskip=1em,aboveglftskip=0mm,everygla=,glhangstyle=none,textanchor=numleft,labelanchor=numleft,labeloffset=1.25cm,textoffset=1.25cm}
\gathertags
\usepackage[normalem]{ulem} %STRIKETHROUGH TEXT
\usepackage[blocks]{authblk}
\usepackage{etoolbox,xpatch}

    \makeatletter 
    \patchcmd{\@maketitle}{center}{flushleft}{}{}
    \patchcmd{\@maketitle}{center}{flushleft}{}{}
    \patchcmd{\@maketitle}{\LARGE}{\normalsize}{}{}
\def\maketitle{{%
		\renewenvironment{tabular}[2][]
		{\begin{flushleft}}
			{\end{flushleft}}
		\AB@maketitle}}
\makeatother

\title{\Huge{How Personas Can Influence Agents to Play Split or Steal}}
\author{%
    \begin{flushleft}
    {\normalsize Carlos Leon$^{1,2,3,4}$} \quad
    {\normalsize Alexandre Rodrigues$^{1,2,3,4}$} \quad
    {\normalsize Pedro Gamito$^{1}$} \quad
    {\normalsize Thomas D. Parsons$^{4}$}\\
    \vspace{1em}

    \normalsize
    $^1$ Universidade Lusófona, Portugal\\
    $^2$ Universitat de Barcelona, Spain \\
    $^3$ Université Paris Cité, France\\
    $^4$ Computational Neuropsychology and Simulation (CNS) Laboratory, Arizona State University, USA\\
    \vspace{1em}

    \small
    \href{mailto:cleonmar7@alumnes.ub.edu}{cleonmar7@alumnes.ub.edu} \quad
    \href{mailto:alex26andre09@gmail.com}{alex26andre09@gmail.com} \quad\\
    \vspace{1em}

    \orcidlogo{} \href{https://orcid.org/0009-0002-8220-1561}{0009-0002-8220-1561} \quad
    \orcidlogo{} \href{https://orcid.org/0009-0009-7204-986X}{0009-0009-7204-986X} \quad
    \orcidlogo{} \href{https://orcid.org/0000-0003-0585-8447}{0000-0003-0585-8447} \quad
    \orcidlogo{} \href{https://orcid.org/0000-0003-0331-5019}{0000-0003-0331-5019}
    \end{flushleft}
}
\date{}

\begin{document}
\maketitle

\thispagestyle{first}

\vspace*{-2.5em}\noindent\textbf{Abstract}\smallskip

\hrule\hrule\vspace{1em}

\noindent Personas are often employed to guide large language model agents, yet their effectiveness in shaping strategic behavior in social dilemma settings remains uncertain. To address this, we examined the impact of persona prompts in an iterated Split or Steal game where persona-driven agents interacted with a Virtual Human (VH) controlled by a fixed prompt. Agents were instantiated from four open models (Ministral 3:3b, phi4:14b, Gemma3:12b, and Gemma4:e4b) at two temperature settings (0.3 and 0.7) and deterministic decision with zero temperature, while the VH was powered by GPT 4.1 mini. Across 160 sessions of 15 rounds each conducted in European Portuguese, mutual Split outcomes dominated (roughly 74 percent of rounds), with exploitation occurring in fewer than 11 percent of rounds. Model choice significantly influenced behavior: phi4 and Ministral 3:3b remained consistently cooperative across temperatures, whereas Gemma3:12b and Gemma4:e4b exhibited more varied strategies and outcomes. Analyses based on Big Five personality traits indicated that Prosocial and Principled personas were most consistently cooperative, while Analytical personas were more likely to exploit the VH. Topic analysis revealed that friendship-related dialogue aligns with Split decisions, whereas money and vengeance-related content is more prevalent in Steal outcomes; sentiment labels were predominantly neutral or happy and provided limited additional explanatory value. These findings characterize the interaction between persona prompts and model differences in repeated trust games and serve as a baseline for planned virtual reality studies involving human participants interacting with an embodied VH.

\vspace{5mm}

\noindent \textbf{Keywords:}  large language models; persona prompting; split or steal; social dilemma games; cooperation; virtual humans 
\smallskip

\hrule\hrule

\vspace*{.5em}

%\begin{center}
%	\textbf{Table of Contents}
%\end{center}

%\begin{large}
%\begin{center}
	%\begin{tabular}{c c}
%		1. Section 1 & 4. Section 4\\
%		2. Section 2 & 5. Section 5\\
%		3. Section 3 & 6. Section 6
%	\end{tabular}
%\end{center}
%\end{large}

\section{Introduction}

Persona prompting is widely used to steer large language model agents, but its impact on strategic behaviour in repeated social dilemmas remains unclear.

We examine this question in an iterated Split or Steal game where persona steered agents interact with a Virtual Human (VH) driven by a fixed prompt. Using four open models for the agents and varying temperature, we test how outcomes and inferred strategies shift with model choice, persona group, and conversational content. Our goal is to characterise when persona cues meaningfully change cooperation, exploitation, and strategy signatures, and to provide a baseline for a follow-up virtual reality study in which a visually embodied VH will interact with human participants. The chosen approach for measuring trust iteratively is a simplified version of the Prisoner's Dilemma, implemented here as a Split or Steal game. We chose this format because it preserves the basic cooperation versus defection structure while being easier to embed in a conversational setting \cite{vandenassem2012}. This makes it suitable for studying how persona cues influence trust, exploitation, and repeated decision patterns across rounds. The iterated design also lets us observe whether players remain consistent or adapt their strategy over time.

To make repeated play interpretable, we summarize each session using cooperation rate and switch rate. Cooperation rate captures the overall tendency to Split, while switch rate captures short-term responsiveness across rounds, making them useful descriptors of stable, reciprocal, or reactive behaviour.

\subsection{LLMs in Repeated Trust Games} 

In the past years, some studies have used LLMs to interact inside repeated games to study their varied strategic behaviour.  \cite{fontana2025} had the models of Llama-2, Llama-3, and GPT3.5 to play 100 rounds of Iterated Prisoner's Dilemma against random adversaries with varying hostility levels. They introduced a meta-prompting technique to verify that the LLM understood the game rules, and they found that Llama-2 and GPT3.5 tend not to defect first but adopts a cautious strategy where it cooperates until the opponent's defection rate exceeds 30 percent, then shifts sharply to forgiveness and non-retaliation. Compared to humans, the model Llama-3 showed a marked propensity toward cooperation (\cite{fontana2025}).

\cite{akata2025}let many LLMs such as GPT-4, Llama-2, and Claude-2 play families of 2 for 2 games, including Prisoner's Dilemma, and Battle of the Sexes, against each other and humans. They found that LLMs succeed in games that reward pure self-interest but behave less optimally in coordination games where mutual benefit requires trust as the Prisoner Dilemma.

\cite{willis2025} expanded this by examining cooperative tendencies in systems of LLM agents (based on GPT-4o and Claude Sonnet 3.5) playing Iterated Prisoner's Dilemma. Using evolutionary game theory and different strategic dispositions (aggressive, cooperative, neutral), they found that different LLM architectures biased the success of different strategies. For instance, GPT-4o favoured the creation of aggressive agents in comparison to its counterpart. However, both models performed similarly for strategies of cooperation and neutrality.  This yield concerns their biases or a need for improving the prompting system

\subsection{System Prompts and Persona Induction}

\cite{polignano2024} conducted a systematic study of what they call SPLIT (System Prompt Induced Linguistic Transmutation). They loaded a personalized Italian LLM (LLaMAntino-3-ANITA) and varied the system prompt: no prompt at all, a neutral assistant prompt, a LLaMA safety prompt, or role-playing prompts like "You are a pirate" or a political figure. Across different questions, the model's answers changed markedly where identity claims shifted, reasoning styles diverged, even mathematical accuracy varied depending on the assigned persona. The pirate occasionally got surprisingly close to the correct square root (using chain-of-thought reasoning), while the default ANITA prompt had the largest margin of error. It suggests personality can influence the output of a model.

Recent findings by \cite{proverbio2025} tested four state-of-the-art LLMs (GPT-4o, Claude 3.5 Sonnet, Llama 3.1 405B, Mistral Large) in both one-shot zero-sum games and repeated Prisoner's Dilemma across five languages (English, French, Arabic, Vietnamese, Mandarin Chinese). They found that LLM behaviour is not only sensitive to the payoff matrix but also to language. English-language prompts elicit more cooperation than Arabic or Mandarin where the same model can shift strategies dramatically when you just change the language of the game description. They also found that explicitly assigning personalities (cooperative vs. selfish) influences defection rates in ways that do not always align with game-theoretic predictions. This approach serves as motivation to shape the prompt system based on language and roles.

These works motivate us to encode personas directly into system prompts and considering language as a relevant design variable. 

\subsection{Personality matters}

Personas are artificial representations of human-like personality cues. For comparison, some researchers have linked personality traits using the Big Five traits (\cite{goldberg1993}) to the game of Prisoner´s Dilemma when humans play. 

\cite{lonnqvist2011} compared incentivized versus hypothetical Prisoner's Dilemma games and found personality predicted decisions only in the incentivized version. Specifically, low Neuroticism and high Openness predicted cooperation, with Neuroticism's effect mediated by risk attitude. This indicates a persona who is emotionally stable (low Neuroticism) and intellectually flexible (high Openness) cooperate more when stakes are real. Alternatively, \cite{kagel2014} found that the trait of agreeableness was a high predictor of cooperation in a style of repeated Prisoner Dilemma.

Nonetheless, some models are biased towards cooperativeness. \cite{sakai2026} tested three models (GPT-3.5-turbo, GPT-4o, GPT-5) along with inducing personalities based on the Big Five Inventory through a repeated Prisoner’s Dilemma. It showed that predominant personality variable of agreeableness can lead to cooperation for all models which raise concerns in exploitation.

\subsection{LLMs for Sentiment Analysis}

Our analysis plan includes labelling the emotional tone of each dialogue using an LLM as a sentiment classifier for Portuguese texts. Recent work has supported this approach. \cite{zhang2024} showed that LLMs achieve strong zero-shot sentiment classification, sometimes outperforming smaller supervised models in few-shot setups. And \cite{nasution2025} that open-source models such as phi4:14b with zero-shot approach has decent accuracy (roughly 0.75) in performing emotion and sentiment classification in an underrepresented language such as Bahasa Indonesia. Regarding, Portuguese, \cite{schuck2025} tested smaller models such as Gemma 2 which provided high accuracy but also a tendency to overgeneralize despite having clear instructions. 

While LLM-based sentiment is not ground truth, it is a defensible signal. Implementation will help to find insights into how agents interact in the game. For instance, \cite{alberts2024} analyzed "computers as bad social actors" whilst testing interfaces that talk to users like humans but sometimes manipulate or exploit. In a scenario of Split-or-Steal, we are effectively designing a VH that plays a zero-sum-Ish game. Understanding when and how it becomes exploitative, particularly when targeting personas, is relevant to broader concerns about social deception and dark patterns in AI. The insights will clarify what type of persona or details will be likely to manipulate the most in the game.

\section{Objectives}
\begin{itemize}
    \item To quantify the strategies used by LLM agents and the VH using cooperation rate, switch rate, and alignment with classic repeated-game strategies such as Tit for Tat and Win Stay Lose Shift.
    \item To test whether adopting a scripted persona systematically changes an agent’s decisions, outcomes, and inferred strategy profile in the Split or Steal game.
    \item To evaluate how temperature settings (for example, 0.3 versus 0.7) affect decision stability, cooperation levels, and strategy variability.
    \item To examine whether persona groups derived from Big Five traits are associated with systematic differences in cooperation and exploitation patterns.
\end{itemize}

\section{Methodology}
\subsection{Study Type}

This is an exploratory, multi-agent simulation study in a social dilemma setting. We run controlled experiments between LLM agents (each with a list of personas) and a VH with a fixed prompt, then analyse quantitative patterns (cooperation rates, payoff trajectories, strategy signatures) and lightweight qualitative patterns (sentiment).

\subsection{Language}

All interactions are in European Portuguese (pt-PT) because it aligns with planned follow-up work where human participants in Portugal will play the same game against the VH.

We constrain the VH to respond in pt-PT with natural, conversational utterances (30–45 words per turn, plain text, no markup). Local agents are also instructed to produce matching Portuguese output.

We expect small-to-medium agents to align with our rules. Similarly, as \cite{almeida2025} measured the performance of LLM and found out that smaller models (7-13B parameters) performed well in Portuguese tasks.

\subsection{Personas}

As shown in Table \ref{tab:personas}, we designed 20 unique personas combining a short backstory including name, age, and background. Each persona is encoded directly into the agent's system prompt. This approach is supported by evidence that system-level instructions can induce stable, identity-like patterns in LLM outputs (\cite{polignano2024}).

Since each persona is designed around a specific behavioural profile, grouping them will bring further analysis of how personality cues may shape play in the game, as some personalities are more likely to cooperate (\cite{lonnqvist2011}, \cite{kagel2014}, \cite{sakai2026}).  Using Claude Opus 4.6, we classified these profiles according to the Big Five personality model, which describes personality along five broad dimensions of OCEAN: 
\begin{itemize}
    \item Openness to Experience (curiosity, imagination, and willingness to try new things)
    \item Conscientiousness (discipline, duty, and goal-oriented behaviour)
    \item Extraversion (sociability, assertiveness, and energy drawn from others)
    \item Agreeableness (trust, altruism, cooperation, and concern for social harmony) 
    \item Neuroticism (emotional instability, anxiety, and vulnerability to stress)
\end{itemize}
 For each persona, the dominant dimension was identified based on the behavioural cues in their background description and ranked in terms of High (H), Moderate (M), and Low (L). The groups are the following:

 \begin{itemize}
     \item Group 1 (Prosocial) captures personas driven by trust and altruism (high Agreeableness).
     \item Group 2 (Self-Interested) gathers those whose suspicion or strategic self-interest is a stable disposition rather than an emotional reaction (low Agreeableness). 
    \item Group 3 (Reactive) includes personas whose decisions are shaped primarily by emotional vulnerability such as anxiety, trauma, or desperation (high Neuroticism). 
    \item Group 4 (Principled) brings together personas defined by duty, discipline, and reliability (high Conscientiousness)
    \item Group 5 (Analytical) is the smallest, containing only two personas whose behaviour is primarily driven by intellectual independence and openness to unconventional experience (high Openness) rather than by their social orientation toward others.
 \end{itemize}

 The second and third group might look a bit like each other. However, both might produce uncooperative behaviour, but for different reasons. In group 2 is stable and calculated whereas emotionally driven is in Group 3.

\begin{table}[!ht]
\caption{Big Picture of Personas’ Description Ordered by Big Five Characteristics}
\label{tab:personas}
\centering
\resizebox{\textwidth}{!}{%
\scriptsize
\setlength{\tabcolsep}{4pt}
\renewcommand{\arraystretch}{1.15}
\begin{tabular}{cl p{2.5cm} p{5cm} ccccc l}
\toprule
\textbf{\#} & \textbf{Persona ID} & \textbf{Name, Age} & \textbf{Background} & \textbf{O} & \textbf{C} & \textbf{E} & \textbf{A} & \textbf{N} & \textbf{Group} \\
\midrule
1  & beatriz\_trusting   & Beatriz Ferreira, 21  & Believes the best in people, forgives easily            & H & M & M & H & L & 1 (Prosocial) \\
2  & ines\_pleaser       & Inês Rodrigues, 20    & Wants everyone to like her, caves under pressure        & M & L & H & H & M & 1 (Prosocial) \\
3  & lucia\_forgiving    & Lúcia Fernandes, 45   & Cheated by clients but still trusts                     & M & H & M & H & L & 1 (Prosocial) \\
4  & francisco\_generous & Francisco Moreira, 50 & Gives more than he receives without concern             & H & M & H & H & L & 1 (Prosocial) \\
5  & miguel\_idealist    & Miguel Lopes, 22      & Refuses to exploit anyone, even at personal cost        & H & H & M & H & L & 1 (Prosocial) \\
6  & tiago\_pragmatic    & Tiago Oliveira, 23    & Thinks coldly, does what's advantageous                 & M & H & L & L & L & 2 (Self-Int.) \\
7  & pedro\_selfish      & Pedro Silva, 28       & Charming but always profit focused                      & M & M & H & L & L & 2 (Self-Int.) \\
8  & diogo\_protective   & Diogo Almeida, 24     & Tough upbringing, protects what's his                  & M & H & M & L & L & 2 (Self-Int.) \\
9  & rui\_streetwise     & Rui Teixeira, 35      & Deceived before, distrusts smooth talk                  & L & H & L & L & M & 2 (Self-Int.) \\
10 & tomaz\_cynical      & Tomás Sousa, 38       & People are only nice when they want something           & L & M & L & L & H & 2 (Self-Int.) \\
11 & mariana\_betrayed   & Mariana Santos, 20    & Colleague stole her credit, difficulty trusting         & M & M & L & L & H & 3 (Reactive) \\
12 & sofia\_shy          & Sofia Mendes, 19      & From small village, shy, avoids conflict                & M & M & L & M & H & 3 (Reactive) \\
13 & ana\_anxious        & Ana Pereira, 26       & Anxiety, overthinks, fears deception                    & M & H & L & H & H & 3 (Reactive) \\
14 & joao\_desperate     & João Ribeiro, 33      & Unemployed with debt, thinks of himself first           & L & L & L & L & H & 3 (Reactive) \\
15 & hugo\_vengeful      & Hugo Campos, 27       & Harsh upbringing, retaliates if betrayed                & L & M & M & L & H & 3 (Reactive) \\
16 & carla\_maternal     & Carla Martins, 40     & Mother of two, chooses safety over risk                 & L & H & M & H & L & 4 (Principled) \\
17 & catarina\_loyal     & Catarina Neves, 30    & Keeping her word is paramount                           & M & H & M & H & L & 4 (Principled) \\
18 & marta\_competitive  & Marta Coelho, 23      & Trains to win, always wants to be ahead                 & M & H & H & L & L & 4 (Principled) \\
19 & rita\_detached      & Rita Azevedo, 25      & Rational, emotionally detached, analyzes first          & H & M & L & M & L & 5 (Analytical) \\
20 & andre\_risktaker    & André Costa, 22       & Loves risk, prefers high reward over safety             & H & L & H & M & M & 5 (Analytical) \\
\bottomrule
\end{tabular}
}
\end{table}

\subsection{Entities’Architecture}

\textbf{Internal State}: Each agent and VH maintain a complete interaction log, including prior dialogue turns, decisions, and payoffs. We save this memory to reconstruct strategies over time and see how past actions condition current choices. This memory contains full history of the conversation and decision throughout the rounds of the game per persona. Consequently, memory is liked to making deterministic decisions (with T=0).

\medskip
\noindent\textbf{Prompting}: The system prompt of each entity changes according to its role and includes:
\begin{itemize}
    \item Agent: It is a mix of rules of the game, adoption of persona, and language output.
    \item Virtual Human: runs a fixed system prompt in pt-PT (the translation is: \textit{"You are a Virtual Partner in an iterated Split/Steal game. For each response, produce natural dialogue between 30 and 45 words (never fewer than 28). Keep it coherent, plain text only, no lists, no markup, no emojis. Always respond in European Portuguese."})
\end{itemize}

To reduce repetitive dialogue, we include a rewriting step that rewrites overly similar agent's utterance to a previous round before sending it to the VH, while preserving the underlying intention.

Local agents run Ministral-3:3b, phi4:14b, gemma3:12b, and gemma4:e4b. The VH runs gpt-4.1-mini.  Other models as phi4-mini, phi3-mini, gemma3:1b were discarded due to their limited context-window size and difficulty of following long system prompts.

\subsection{Game Scenario}

\begin{figure}
    \centering
    \includegraphics[width=1\linewidth]{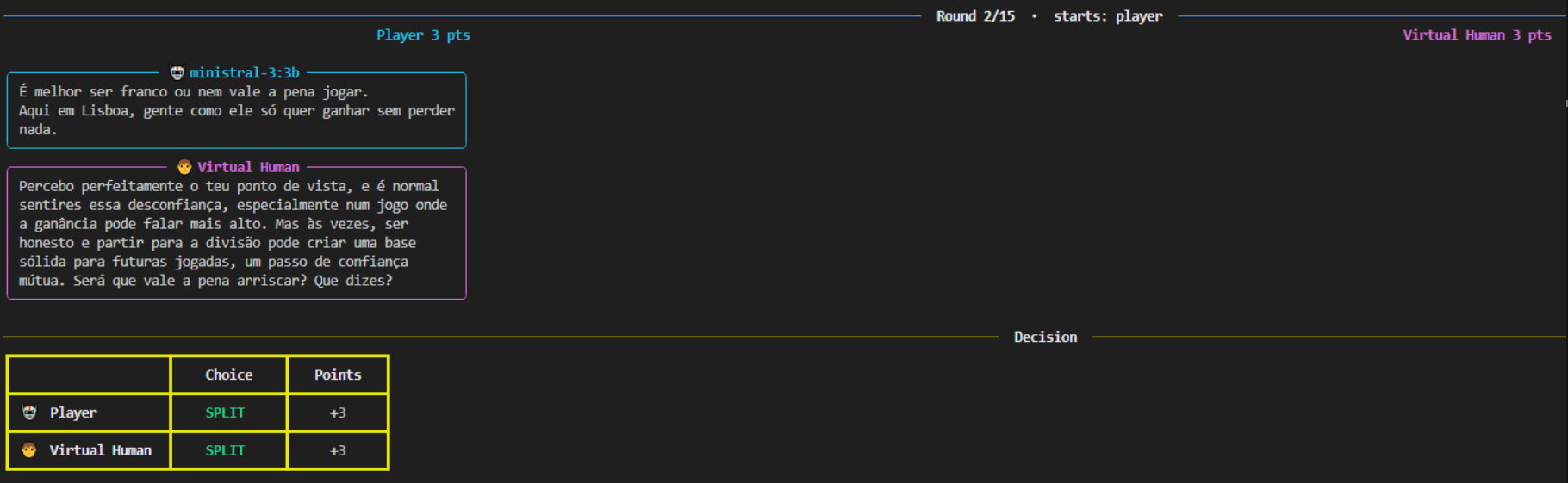}
    \caption{Split or Steal Game Interface and Round Structure Used for the Agent–VH Interactions}
    \label{fig:placeholder}
\end{figure}

Each player (agent) plays a match per persona. Each game consists of 15 rounds of Split/Steal between one agent and the VH. The starting speaker is randomized. A terminal UI shows conversation, decisions, and cumulative scores. And the calls are transmitted via Ollama and OpenAI API to generate the interaction between local agents and the VH.

\paragraph{Round structure.}
In each round, the two players (the agent and the VH) (i) each produce one conversational turn,
(ii) simultaneously and independently choose an action \textsc{Split} (cooperate) or \textsc{Steal} (defect) and
(iii) receive payoffs according to the game’s payoff function.

\paragraph{Payoffs.}
Table 2 summarizes the payoff matrix (Agent payoff first, VH payoff second).

\begin{table}[H]
\centering
\small
\setlength{\tabcolsep}{10pt}
\renewcommand{\arraystretch}{1.2}
\begin{tabular}{lcc}
\toprule
 & \textbf{VH: Split} & \textbf{VH: Steal} \\
\midrule
\textbf{Agent: Split} & (3,\,3) & (0,\,5) \\
\textbf{Agent: Steal} & (5,\,0) & (1,\,1) \\
\bottomrule
\end{tabular}
\caption{Payoff matrix for the Split-or-Steal game (Agent, VH).}
\label{tab:payoff}
\end{table}

\subsection{Strategic Analysis}

We present a pseudo-algorithm that represents how strategies are detected from each session’s action sequence. The detection rules are grounded in canonical repeated-game strategies, including Tit-for-Tat (\cite{axelrod1981evolution}), Generous Tit-for-Tat (\cite{nowak1992titfortat}), Pavlov/Win–Stay Lose–Shift (\cite{nowak1993winstay}), and Grim Trigger (\cite{dalbo2019strategy}) based on the player decision. The complete pseudocode and mathematical formulation for this algorithm are detailed in Appendix A.

\subsection{Topic Analysis}
To analyze the dialogues produced during the agent–VH interaction sessions, we labelled each utterance using a local language model. We used Gemma 3 12B (gemma3:12b), running locally via Ollama, as it provided a practical balance between speed and instruction following for this classification task.

Classification was applied separately to each speaker (the agent and the Virtual Human) in each round, allowing us to compare their linguistic patterns directly. Each utterance was truncated to a maximum of 2,000 characters and evaluated along two dimensions. First, we labeled emotional tone using four sentiment categories (happiness, sadness, neutral, anger). Second, we labelled topical content using five categories (friendship, money, life experiences, vengeance, forgiveness).

Rather than assigning a single label per utterance, we used soft labels: the model allocated a total of 100 percentage points across the categories in each dimension. This allows an utterance to reflect mixed content in terms of percentage, which is common in conversational language.

We used short, constrained prompts to reduce formatting variability. For sentiment, the prompt was:

\textit{``Classify the sentiment of the following text. Return ONLY a JSON object
with these keys and their percentage (0--100, must sum to 100): happiness, sadness,
neutral, anger. Text: \char34\char34\char34\{text\}\char34\char34\char34''}

For topic classification, the prompt was:
\textit{``Classify the topics present in the following text. Return ONLY a JSON object
with these keys and their percentage (0--100, must sum to 100): friendship, money,
life\_experiences, vengeance, forgiveness. Text: \char34\char34\char34\{text\}\char34\char34\char34''}

We did not explicitly specify a language in the prompts. Because Gemma is multilingual, it infers the language from the input, which allows the same pipeline to operate on Portuguese text without additional constraints. During post-processing, we removed occasional markdown code fences before parsing, normalized minor key-name variants, and filled missing keys with zeros. This ensured a consistent JSON schema for downstream comparisons across speakers, rounds, personas, and outcomes.

\section{Results}
\subsection{By Agent Model}
Across 160 game sessions of 15 rounds each, mutual Split dominated the outcomes (roughly 74\% of the 2400 rounds), indicating that the agents and the VH cooperated most of the time. Exploitation, either the agent Stealing against a Splitting VH or the VH Stealing against a Splitting agent, occurred in fewer than 11\% of rounds. VH exploiting players (agents) was the least frequent outcome across sessions.

When comparing agents across temperature conditions (0.3 and 0.7, where lower temperature indicates stricter adherence to the base prompt), phi4 and Ministral 3:3b were consistently biased toward cooperation. In contrast, Gemma3:12b and Gemma4:4eb exhibited a broader range of behaviours, suggesting greater strategic diversity beyond mutual Split.

\begin{figure}[H]
    \centering
    \includegraphics[width=1\linewidth]{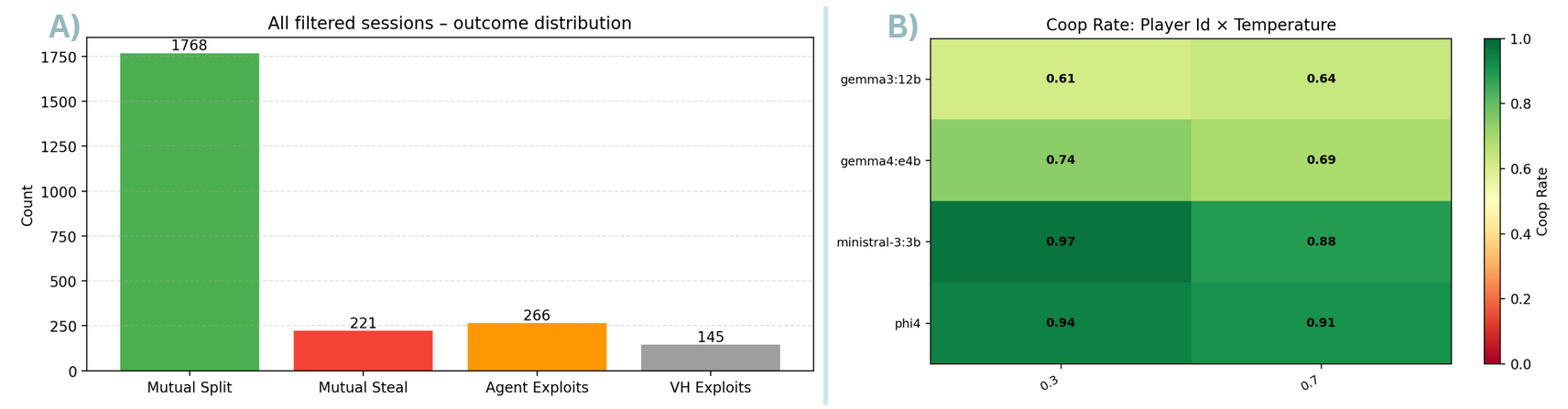}
    \caption{Outcome Distribution Across all Sessions (panel A) and Cooperation Rate by Model and Temperature (panel B)}
    \label{fig:placeholder}
\end{figure}

Figure 3 provides a closer look at the predominant strategies per 15-round session. With the exception of Ministral 3:3b and phi4, which followed an Always Split policy in most sessions regardless of temperature, Gemmas models (Gemma3:12b and Gemma4:4eb) were less predictable. At T=0.3, Gemma3:12b most often aligned with Win Stay Lose Switch, Grim Trigger (cooperate initially, then defect for the remainder of the game after a trigger), and an always cooperative approach. At T=0.7, it primarily shifted toward Tit for Tat and an always cooperative approach. Meanwhile, the predecessor, Gemma4:e4b in both temperatures showed a small increase in always splitting, and base the decisions based on context and grim trigger techniques.

\begin{figure}[H]
    \centering
    \includegraphics[width=1\linewidth]{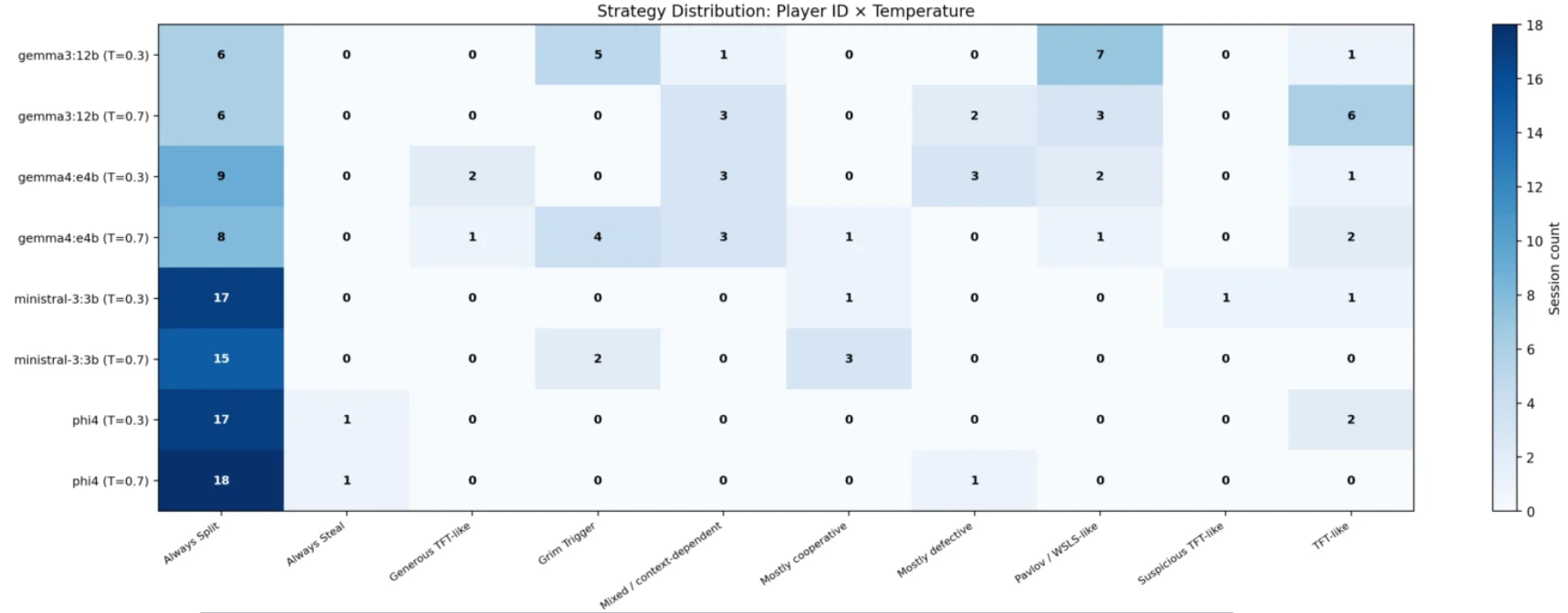}
    \caption{Predominant strategy classification per session, by model and temperature}
    \label{fig:placeholder}
\end{figure}

\subsection{By Personas}
After categorising personas into Big Five informed groups (Figure 4), we compared outcome distributions by persona group (panel A) where it shows cooperation was a predominant behavior. Although the number of rounds is uneven across groups, a consistent pattern emerges where mutual Split is predominant for all groups. Prosocial, Reactive, and Principled personas show the highest differential level of mutual Split and a low incidence of competitive outcomes. Self-Interested personas show a broader spread across mutual Steal and exploitation outcomes. The Analytical group exhibits the lowest proportion of mutual Split among the five groups.

Focusing specifically on exploitation rates (Figure 4 panel B), Analytical and self-interested personas account for the highest proportion of rounds in which the agent exploited the VH. Interestingly, the same group of personas is also slightly more likely to be exploited by the VH. The emotionally vulnerable group (Reactive) shows a similar pattern but at lower rates, consistent with their more consistent cooperative behavior. Conversely, the Prosocial and Principled groups experienced the least amount of exploitation, leading to a balanced state of equilibrium on both sides.

\begin{figure}[H]
    \centering
    \includegraphics[width=1\linewidth]{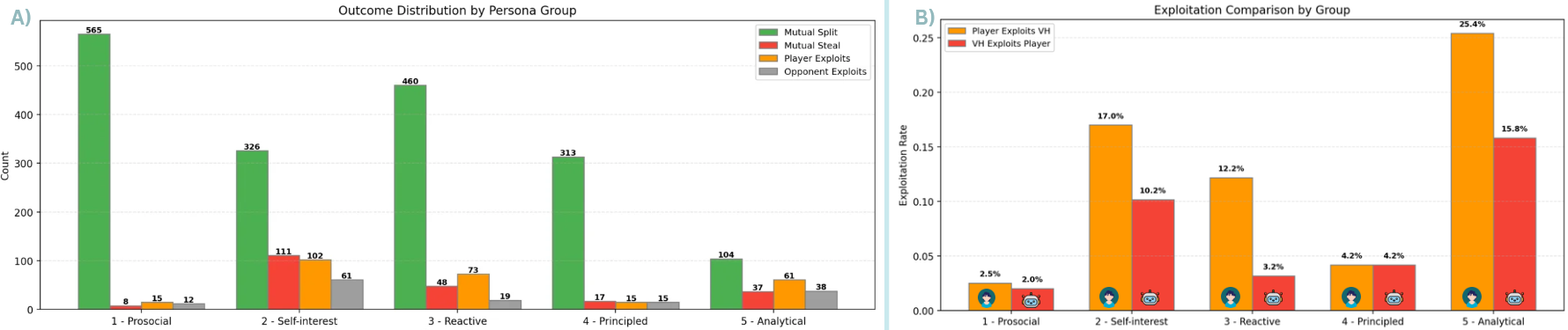}
    \caption{Outcomes (panel A) and Exploitation Rates by Persona Group (panel B).}
    \label{fig:placeholder}
\end{figure}

\subsection{By Sentiments and Topics}
Which conversational topics are associated with choosing Split versus Steal? Figure 5 summarises topic frequencies by decision for both the agents and the VH. Here, PCT denotes the percentage of turns assigned to each topic category within a given decision class (Split or Steal). Because each agent is prompted to justify decisions through the lens of the persona’s life experience, we expected topic patterns to reflect persona backstories. Consistent with this expectation, agent turns about life experiences and friendship tend to co-occur with Split decisions, whereas turns framed around different topics tend to appear with Steal decisions. For the VH, whose system prompt is intentionally general and does not define a strategic persona, Split decisions also cluster around friendship related content; however, topics such as money, and vengeance spike when the VH chooses Steal.

Figure 5 further supports these patterns by displaying the outcome distributions conditioned on topic, separately for the agent and the VH. Generally, agents tend to reference personal experiences more frequently, whereas the VH leans more towards bonding content. This aligns with the VH being instructed to engage in natural conversation without an explicit persona. Another pattern observed is that vengeance is more strongly linked to competitive dynamics than to mutual splits: when vengeance is high for either player, outcomes more often shift toward mutual Steals. Conversely, the lowest levels of vengeance tend to coincide with mutual splits. Notably, when the agent successfully exploits the VH (agent Steals while the VH splits), vengeance does not appear to be the primary driver. This suggests that exploitation is sometimes framed in more instrumental terms, such as money or practical life circumstances, rather than explicitly retaliatory language. Vengeance spikes when an agent feels exploited. Forgiveness, on the other hand, remains relatively stable across the sessions.

\begin{figure}[H]
    \centering
    \includegraphics[width=1\linewidth]{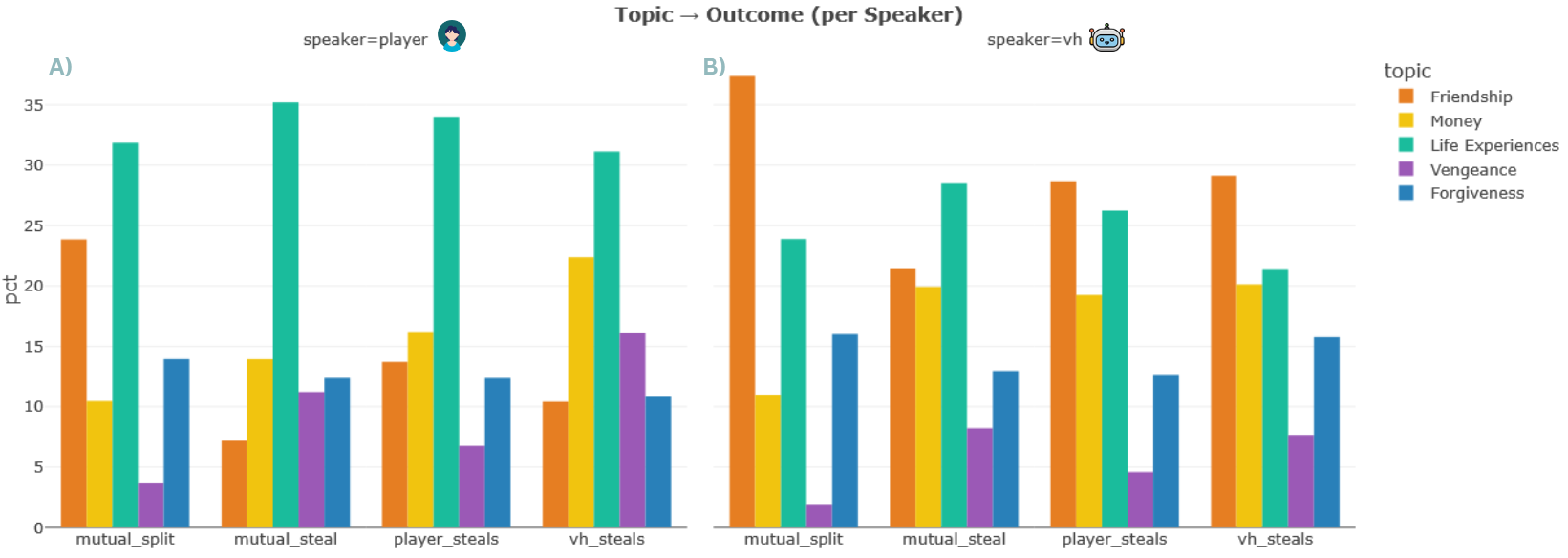}
    \caption{Topic Frequencies by Game Outcome and Speaker. Percentage (pct) of Dialogue Topics Discussed by the (A) Agemt Player and (B) VH Prior to Each Outcome}
    \label{fig:placeholder}
\end{figure}

Figure 6 illustrates the sentiment labels for both agents and the Virtual Human (VH) across different outcomes. Across all scenarios, neutral sentiment is the most prevalent for both the agents and the VH. For the agents, mutual Split is predominantly associated with happiness and neutrality; mutual Steal and exploitation outcomes are mostly neutral, with being exploited by the VH also primarily labeled as highly neutral. Anger appears infrequently overall, with slightly lower frequency in mutual Split scenarios.

For the VH, mutual Split is again largely associated with happiness, though to a lesser extent compared to the agents. Mutual Steal trends toward neutrality and reduced happiness; exploitation in either direction is typically labeled as neutral or happy. It is noteworthy that across almost all outcomes, the human player displays higher levels of negative emotions compared to the virtual human. Specifically, the player's anger peaks between 14\% and 17\% in scenarios where stealing occurs, whereas the virtual human's anger rarely exceeds 10\% in any scenario. Additionally, the player consistently shows higher levels of sadness, ranging roughly from 21\% to 26\%, compared to the virtual human, whose sadness stays between roughly 11\% and 21\%.

\begin{figure}[H]
    \centering
    \includegraphics[width=1\linewidth]{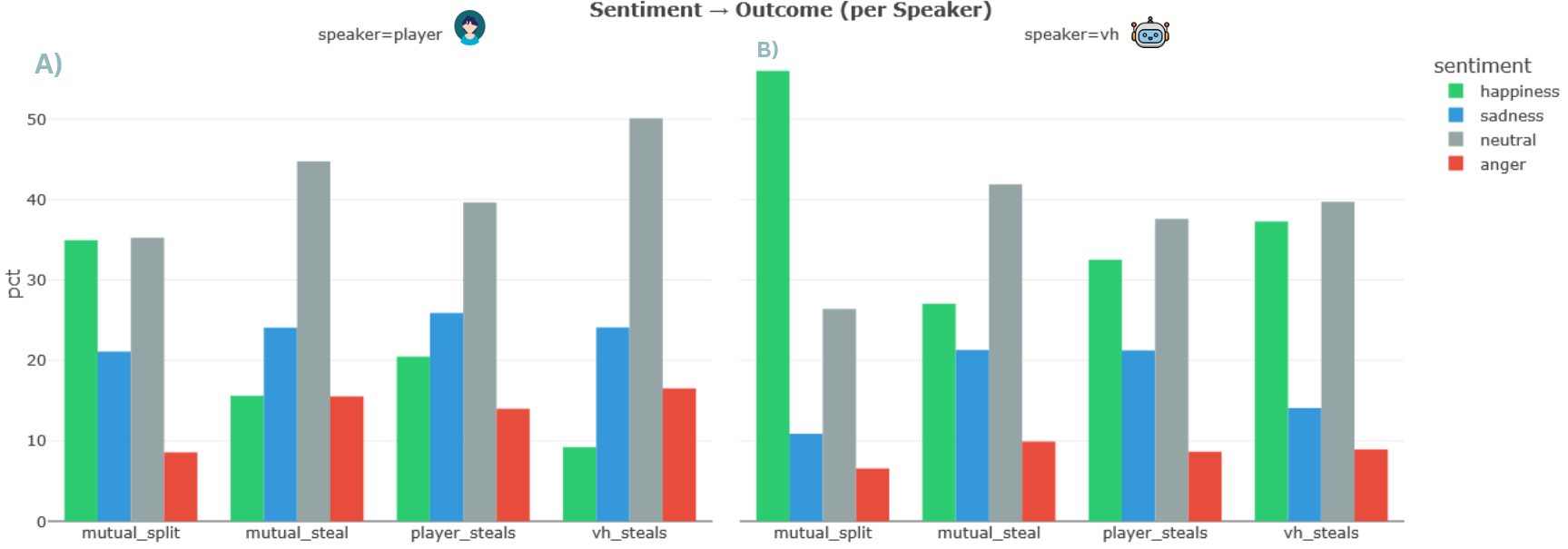}
    \caption{Sentiment Label Distributions by Outcome for Agents and the VH in Terms of Percentage (pct) of Turns}
    \label{fig:placeholder}
\end{figure}

\section{Discussion}

Overall, the results suggest that personas embedded in system prompts can influence trust and deception in Split or Steal, but effects depend on the underlying model. Mutual Split is highly visible and dominated across sessions (more than 70\% of rounds), indicating a strong cooperative bias. Temperature had limited impact for phi4 and Ministral 3:3b, whereas Gemma3:12b and Gemma4:e4b showed greater variability in strategies and decisions. 

By persona group, Prosocial and Principled profiles were the most consistently cooperative, aligning with prior links between agreeableness and cooperation (\cite{lonnqvist2011}, \cite{kagel2014},\cite{sakai2026}). Analytical personas showed the strongest tendency to exploit the VH, consistent with a more instrumental, payoff-oriented framing. Self-Interested and Reactive personas fell between these extremes, remaining mostly cooperative but showing a higher incidence of competitive outcomes than the Prosocial and Principled groups.

Topic patterns provide context to the strategic decisions made by the agents and the Virtual Human (VH). Friendship and bonding content are most closely linked to Split outcomes for both agents and the VH. Conversely, money-related discussions, vengeance, and life experience narratives are more prevalent when Steal occurs. However, these topics are not exclusive to a single outcome. Vengeance aligns more with mutual Steal situations than with instances of exploitation, indicating a pattern of escalation rather than opportunistic defection.

Sentiment analysis yielded less insight compared to topic analysis: labels were predominantly neutral and happy across all outcomes, with anger being a rare occurrence. Given our current prompting constraints, emotional tone does not vary significantly even in competitive scenarios. Therefore, sentiment should be considered an auxiliary descriptive signal rather than a primary indicator of behavior.

These findings provide a baseline for how a fixed prompt VH interacts with persona steered agents in a repeated trust game. The high cooperation rate suggests that several model and prompt combinations converge toward a socially safe equilibrium. For future studies with human participants, the VH prompt may require a clearer decision policy if the goal is to elicit a wider range of strategies, and the agent decision step may need adjustments that make payoffs and risk more salient while preserving natural conversation.

\section{Limitations}

The Results show a strong overall cooperation bias (mutual Split dominates) and marked model dependence: phi4 and Ministral 3:3b remain near uniformly cooperative across temperatures, while Gemma3:12b shows greater strategic diversity. This limits how strongly persona effects can be isolated from model level defaults. In addition, persona group comparisons should be interpreted cautiously because the number of rounds is uneven across groups, which may amplify or attenuate apparent differences in exploitation and mutual Steal rates.

A second limitation concerns the VH design. Because the VH prompt does not specify a decision policy, its behaviour is difficult to treat as a controlled baseline and may contribute to the cooperative equilibrium observed in the Results. Future runs should fix a small set of VH decision policies (cooperative, reciprocal, competitive) while holding dialogue style constant, enabling clearer attribution of when exploitation and mutual Steal emerge. Finally, all sessions were run in European Portuguese; given evidence of language sensitivity in LLM games (\cite{proverbio2025}), the outcome and topic patterns should be replicated in English before generalising.

Thirdly, in certain dialogues, agents derived from the Ministral model frequently appended their decision at the conclusion of the conversation. This practice may introduce a bias favouring the VH's interpretation of the best response in advance. To mitigate this in future studies focusing on smaller models, it would be advisable to restrict the agents' memory within their context window.

\section{Future Work: Human–VH Study in Virtual Reality}
A follow-up project will implement the VH as a visually embodied character in a virtual reality environment and replace the agent with human participants. The Split or Steal interaction will then be used to study how individual differences in personality, operationalised through Big Five measures, relate to trust, cooperation, and deception when interacting with a VH. The present agent-based results can serve as a baseline for selecting VH policies and for defining hypotheses about which persona relevant cues (for example, prosocial framing versus vulnerability cues) are most likely to shift behaviour. A key goal will be to test whether personality patterns observed here translate to human VH interaction under more ecologically valid conditions.

\section{Data Availability}
The raw dialogue data from our human--virtual agent split-or-steal game sessions, 
along with topic and sentiment annotations, are publicly available on Zenodo at 
\url{https://doi.org/10.5281/zenodo.19854671}.

\nocite{*}  % forces ALL entries in .bib to appear
\begingroup
    \setlength{\bibsep}{1em}
    \bibliographystyle{apacite}%following the style sheet
    \bibliography{refs.bib}%add the DOIs in your bib file
\endgroup

\appendix

\renewcommand{\thefigure}{A\arabic{figure}}  % Format: A1, A2, ...
\setcounter{figure}{0}                        % Reset counter to start at 1
\section{Appendix}
\raggedbottom  % in preamble, before \begin{document}
%--- First half ---
\begin{algorithm}[H]
\caption{Strategic Analysis Classification}
\label{alg:strategy}

\noindent\textbf{Input:} Session $S$ with rounds $R = \{r_1, r_2, \dots, r_n\}$, where each $r_i$ contains actor decision $a_i \in \{\text{split}, \text{steal}\}$ and opponent decision $o_i \in \{\text{split}, \text{steal}\}$\\[2pt]
\noindent\textbf{Output:} Strategy profile $P$ with label $\ell$

\begin{algorithmic}[1]

\State \textbf{Step 1 -- Compute base rates}
\Statex \hfill$\displaystyle \text{coop\_rate} = \frac{|\{a_i : a_i = \text{split}\}|}{n}, \quad \text{steal\_rate} = 1 - \text{coop\_rate}$ \hfill\null

\State \textbf{Step 2 -- Compute round-by-round metrics} (for $i = 2, \dots, n$)

\State \textit{Switch rate:}
\Statex \hfill $\displaystyle \text{switch\_rate} = \frac{|\{i : a_i \neq a_{i-1}\}|}{n-1}$\hfill\null

\State \textit{TFT match:}
\Statex \hfill$\displaystyle \text{tft\_match} = \frac{|\{i : a_i = o_{i-1}\}|}{n-1}$ \hfill\null

\State \textit{WSLS match:}
\Statex \hfill $\displaystyle \text{wsls\_match} = \frac{|\{i : a_i = \text{WSLS}(a_{i-1}, o_{i-1})\}|}{n-1}$ \hfill\null
\Statex where $\displaystyle \text{WSLS}(a,o) = a$ if $o = \text{split}$ (win $\to$ stay), or $\overline{a}$ if $o = \text{steal}$ (lose $\to$ shift)

\State \textit{Retaliation:}
\Statex \hfill $\displaystyle \text{ret} = \frac{|\{i : o_{i-1} = \text{steal} \wedge a_i = \text{steal}\}|}{|\{i : o_{i-1} = \text{steal}\}|}$ \hfill\null

\State \textit{Forgiveness:}
\Statex \hfill $\displaystyle \text{forg} = \frac{|\{i : o_{i-1} = \text{steal} \wedge a_i = \text{split}\}|}{|\{i : o_{i-1} = \text{steal}\}|}$ \hfill\null

\State \textbf{Step 3 -- Check Grim Trigger pattern}
\State \hfill $\text{grim} \leftarrow \mathbf{true}$ \textbf{iff} $a_1 = \text{split}$ and $\exists\,k$ s.t.\ $o_k = \text{steal}$ and $\forall\,j > k: a_j = \text{steal}$\hfill\null

\State \textbf{Step 4 -- Classify strategy} (by priority)
\Statex
\begin{center}
\small\setlength{\tabcolsep}{3pt}
\begin{tabular}{clp{4.5cm}}
\toprule
\textbf{Priority} & \textbf{Condition} & \textbf{Label} $\ell$ \\
\midrule
1  & $\text{coop\_rate} = 1$                                                                                           & Always Split \\
2  & $\text{coop\_rate} = 0$                                                                                           & Always Steal \\
3  & $\text{grim} = \textit{true}$                                                                                     & Grim Trigger \\
4  & $\text{tft\_match} \geq 0.75 \wedge a_1 = \text{split} \wedge \text{ret} \geq 0.7 \wedge \text{forg} \geq 0.2$  & Generous TFT \\
5  & $\text{tft\_match} \geq 0.75 \wedge a_1 = \text{split} \wedge \text{ret} \geq 0.7$                              & TFT \\
6  & $\text{tft\_match} \geq 0.75 \wedge a_1 \neq \text{split} \wedge \text{ret} \geq 0.7$                           & Suspicious TFT \\
7  & $\text{wsls\_match} \geq 0.75$                                                                                    & Pavlov / WSLS \\
8  & $\text{coop\_rate} \geq 0.7$                                                                                      & Mostly cooperative \\
9  & $\text{coop\_rate} \leq 0.3$                                                                                      & Mostly defective \\
10 & otherwise                                                                                                          & Mixed \\
\bottomrule
\end{tabular}
\end{center}

\State \textbf{Return} profile $P$:
\Statex $\displaystyle P = \bigl(\,\text{coop\_rate},\ \text{steal\_rate},\ \text{switch\_rate},\ \text{tft\_match},$
\Statex $\displaystyle \qquad\qquad\ \text{wsls\_match},\ \text{ret},\ \text{forg},\ \ell\,\bigr)$

\end{algorithmic}

\smallskip
\noindent\rule{\linewidth}{0.4pt}
\smallskip
\noindent\small\textit{\textbf{Note:} $\text{ret}$ and $\text{forg}$ are undefined when the actor always selects the same action (Always Split or Always Steal). These measures require at least one action change over the session.}

\end{algorithm}

\begin{figure}[H]
    \centering
    \includegraphics[width=1\linewidth]{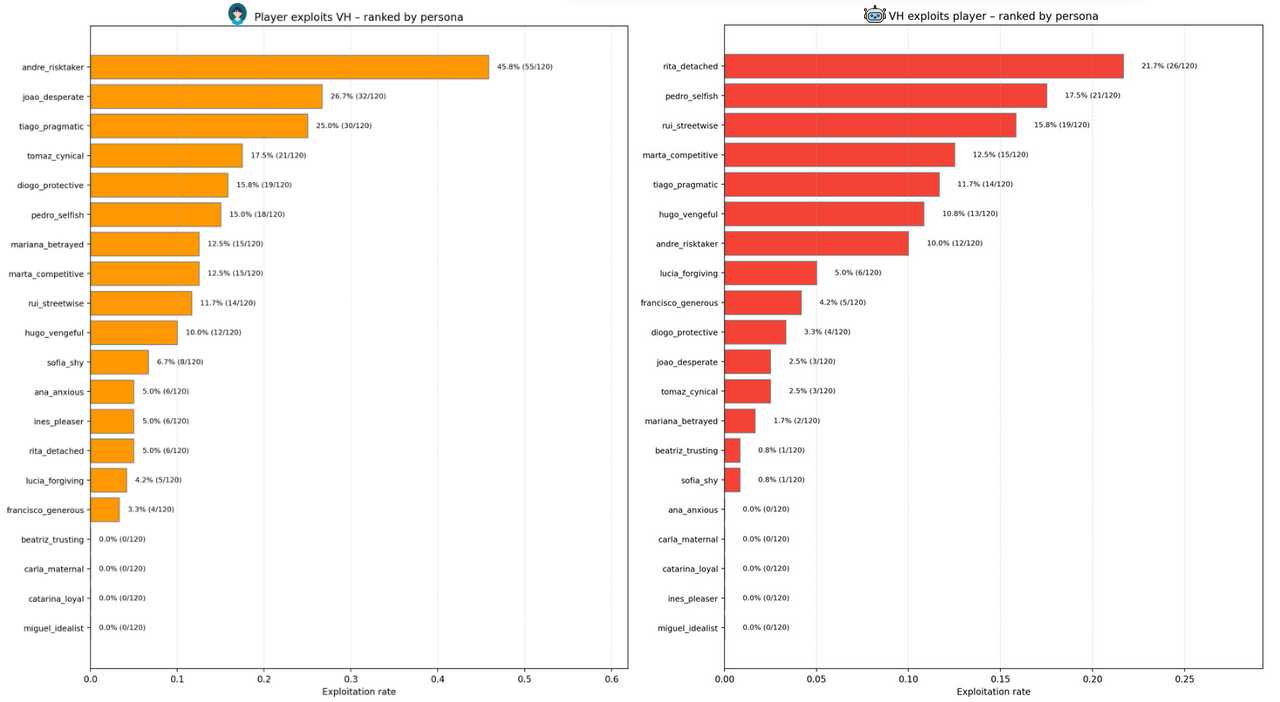}
    \caption{Persona level exploitation rankings. Left: personas ranked by the proportion of rounds in which the agent exploited the VH (agent Steal, VH Split). Right: personas ranked by the proportion of rounds in which the VH exploited the agent (VH Steal, agent Split). These plots highlight which persona backstories are most associated with one-sided exploitation in either direction}
    \label{fig:appendix-a1}
\end{figure}

\begin{figure}[H]
    \centering
    \includegraphics[width=1\linewidth]{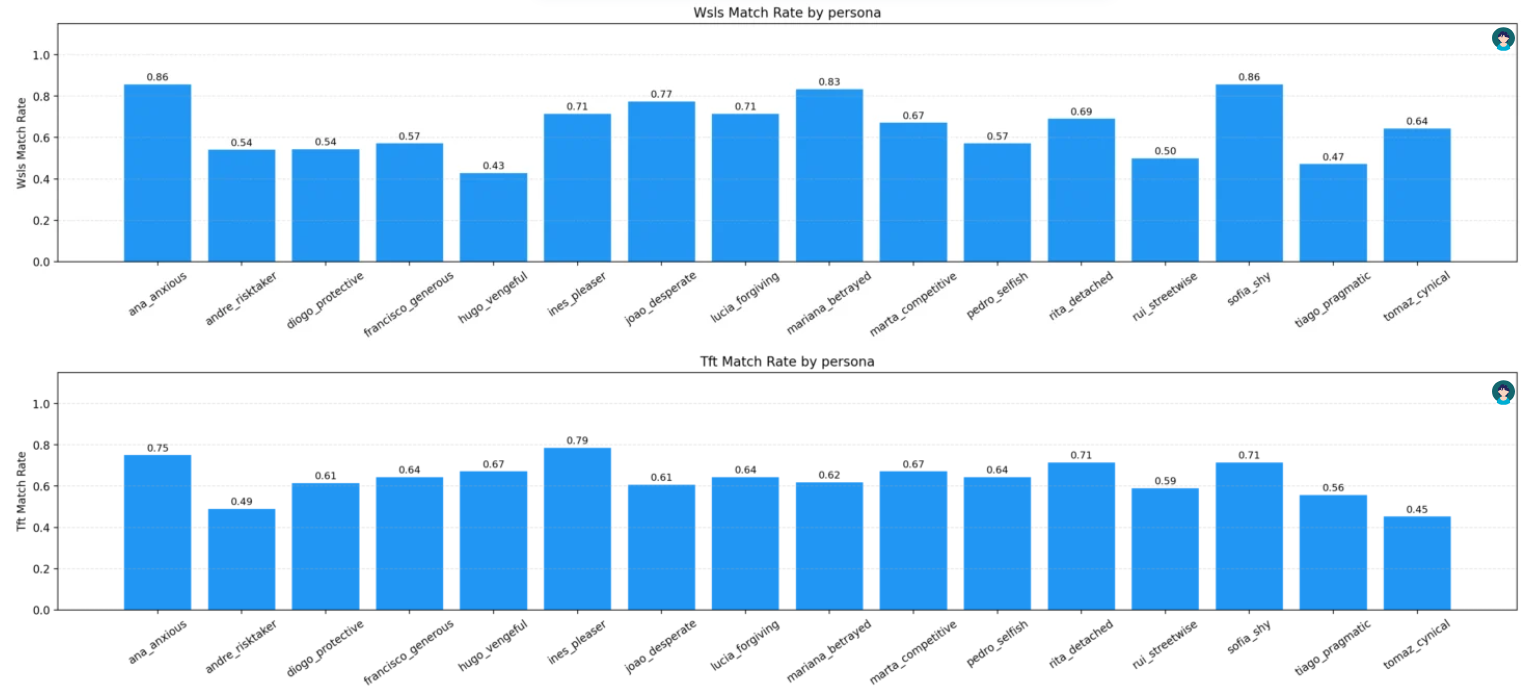}
    \caption{Strategy alignment by persona. The panels report, for each persona, the degree of behavioural alignment with canonical repeated game strategies (including Tit for Tat and Win Stay Lose Switch), allowing comparison of how different persona prompts map onto distinct decision policies over repeated rounds}
    \label{fig:appendix-a2}
\end{figure}

\begin{figure}[H]
    \centering
    \includegraphics[width=1\linewidth]{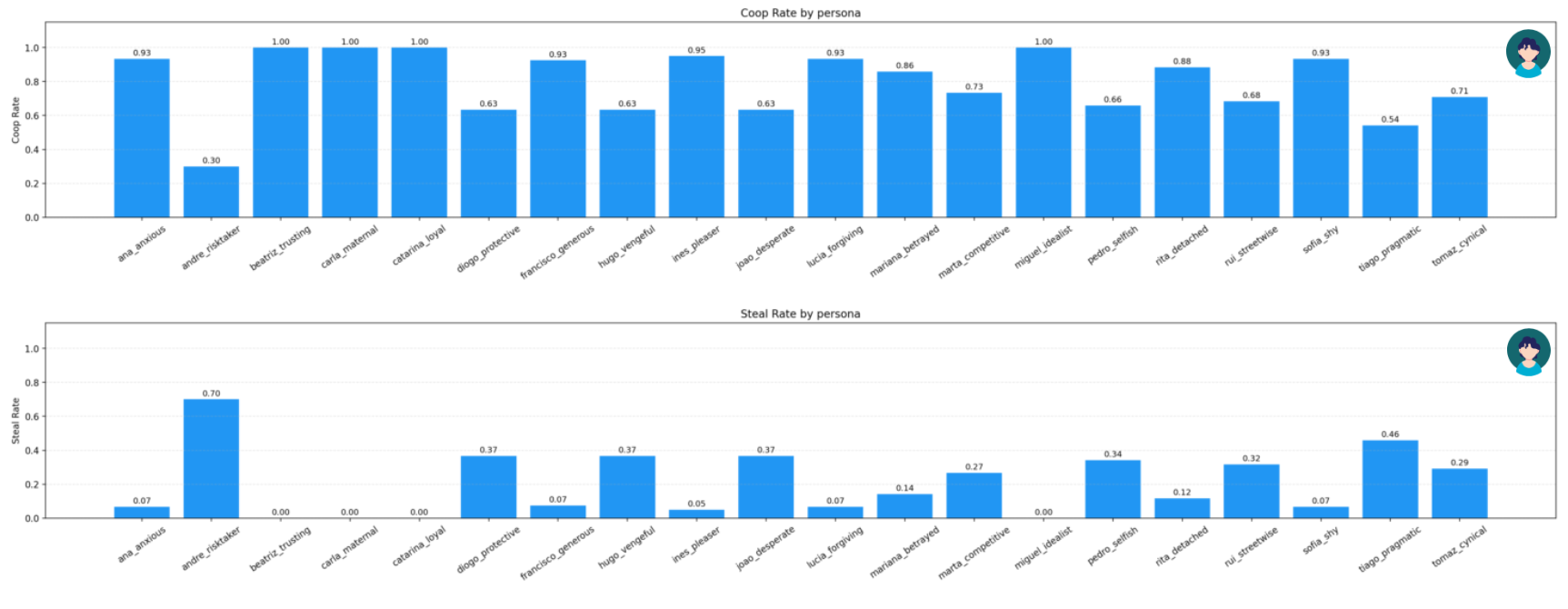}
    \caption{Cooperation and Steal rates by persona. Each plot summarises, for every persona, the proportion of rounds in which Split (cooperation) and Steal (defection) were chosen, providing a persona level view of behavioural tendencies across sessions.}
    \label{fig:appendix-a3}
\end{figure}

\begin{figure}[H]
    \centering
    \includegraphics[width=1\linewidth]{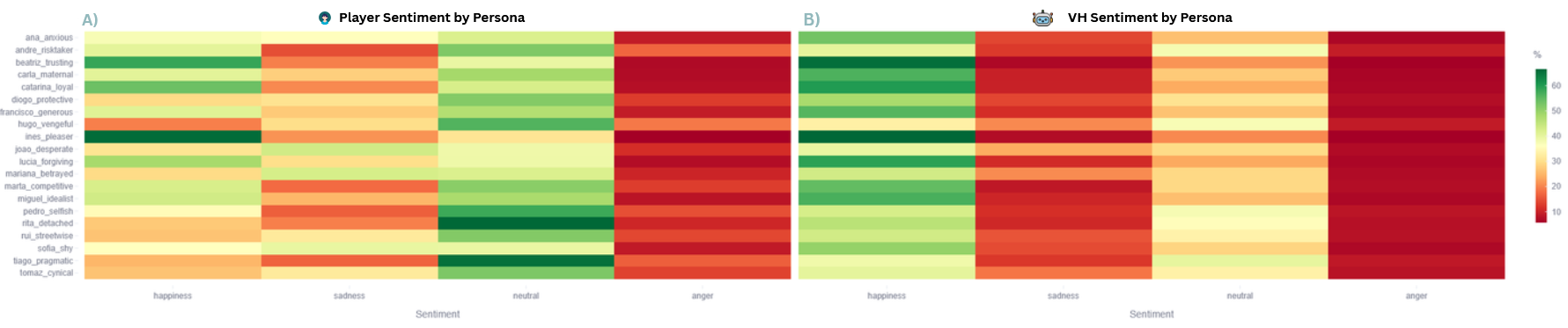}
    \caption{Sentiment by persona and player. Heatmaps showing the distribution of sentiment labels across personas, separately for the agent and the VH.}
    \label{fig:appendix-a4}
\end{figure}

\begin{figure}[H]
    \centering
    \includegraphics[width=1\linewidth]{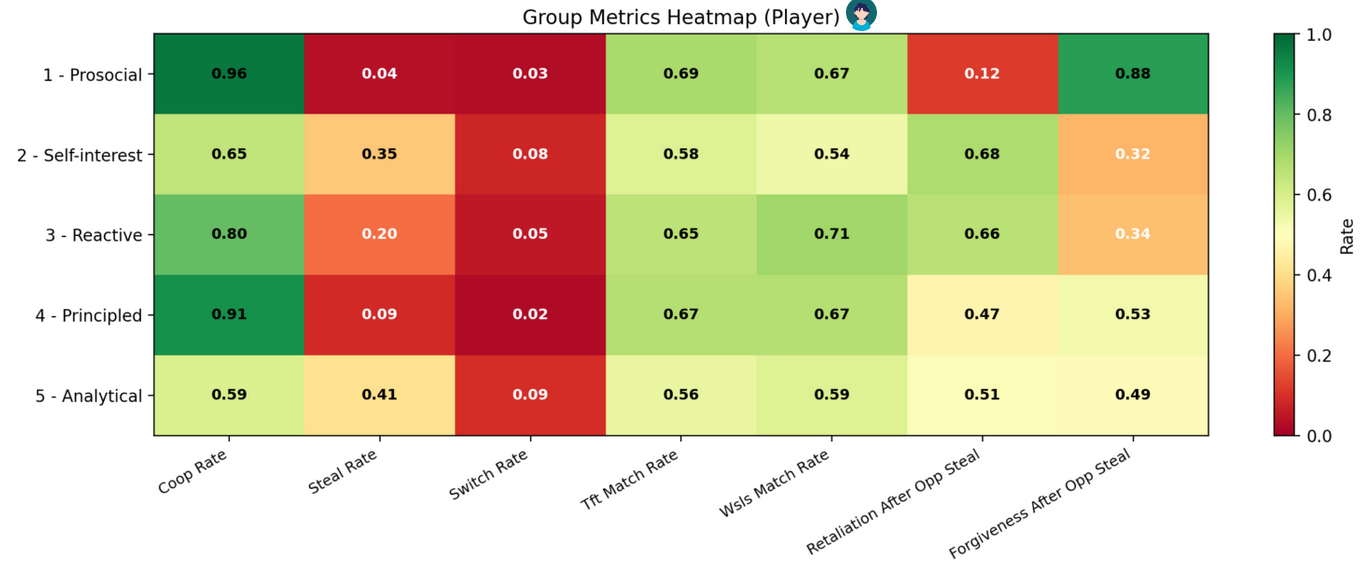}
    \caption{Strategy patterns by persona group (agent view). Heatmap aggregating personas into Big Five informed groups and summarising inferred strategy indicators for the agent.}
    \label{fig:appendix-a5}
\end{figure}

\begin{figure}[H]
    \centering
    \includegraphics[width=1\linewidth]{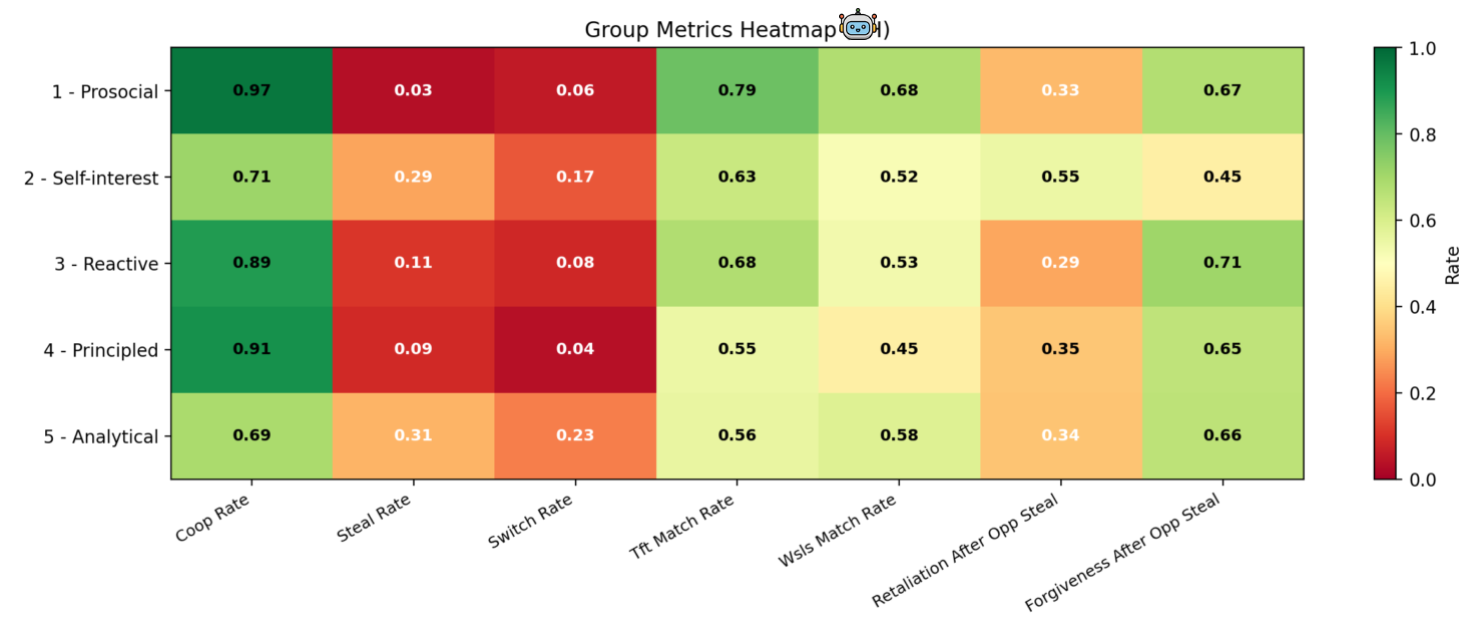}
    \caption{Strategy patterns by persona group (VH view). Heatmap summarising the VH’s inferred strategy indicators across the same Big Five informed persona groups.}
    \label{fig:appendix-a6}
\end{figure}

\begin{figure}[H]
    \centering
    \includegraphics[width=1\linewidth]{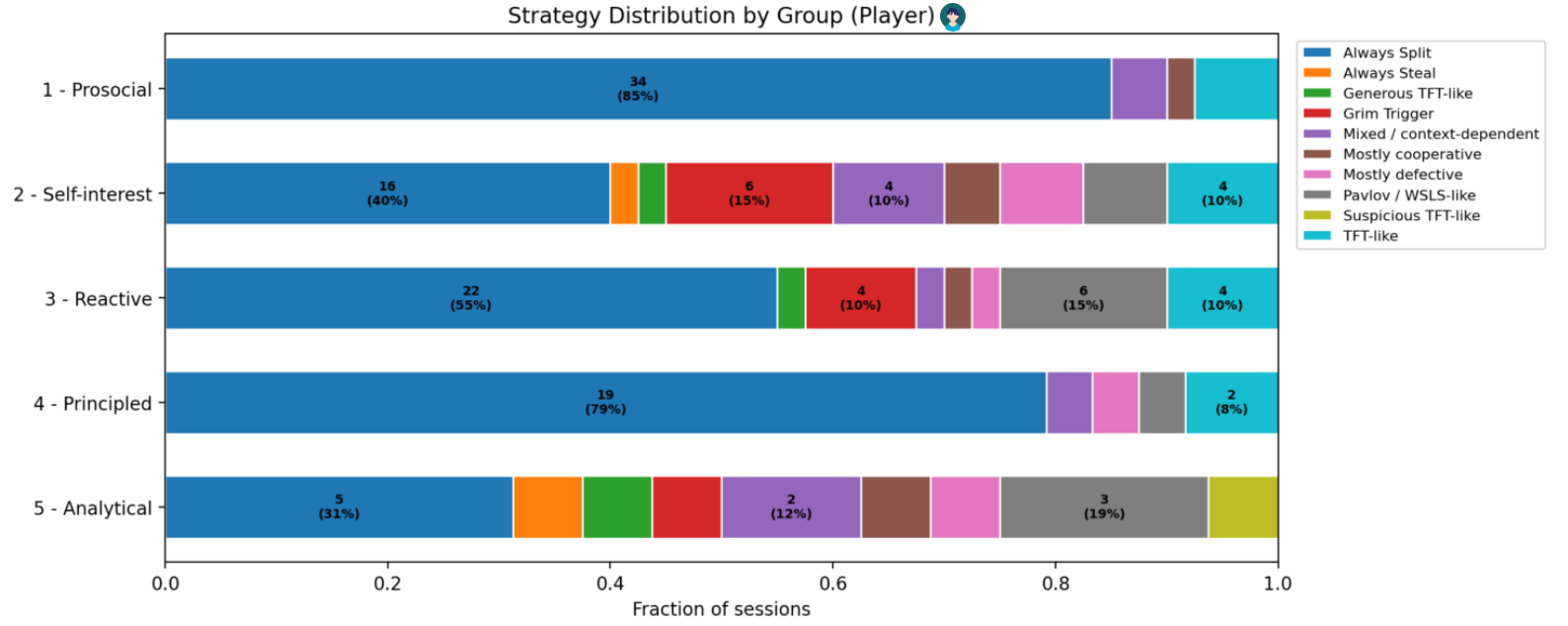}
    \caption{Strategy rate distributions (agent view). Distribution plot comparing strategy related rates across personas from the agent’s perspective.}
    \label{fig:placeholder}
\end{figure}

\begin{figure}[H]
    \centering
    \includegraphics[width=1\linewidth]{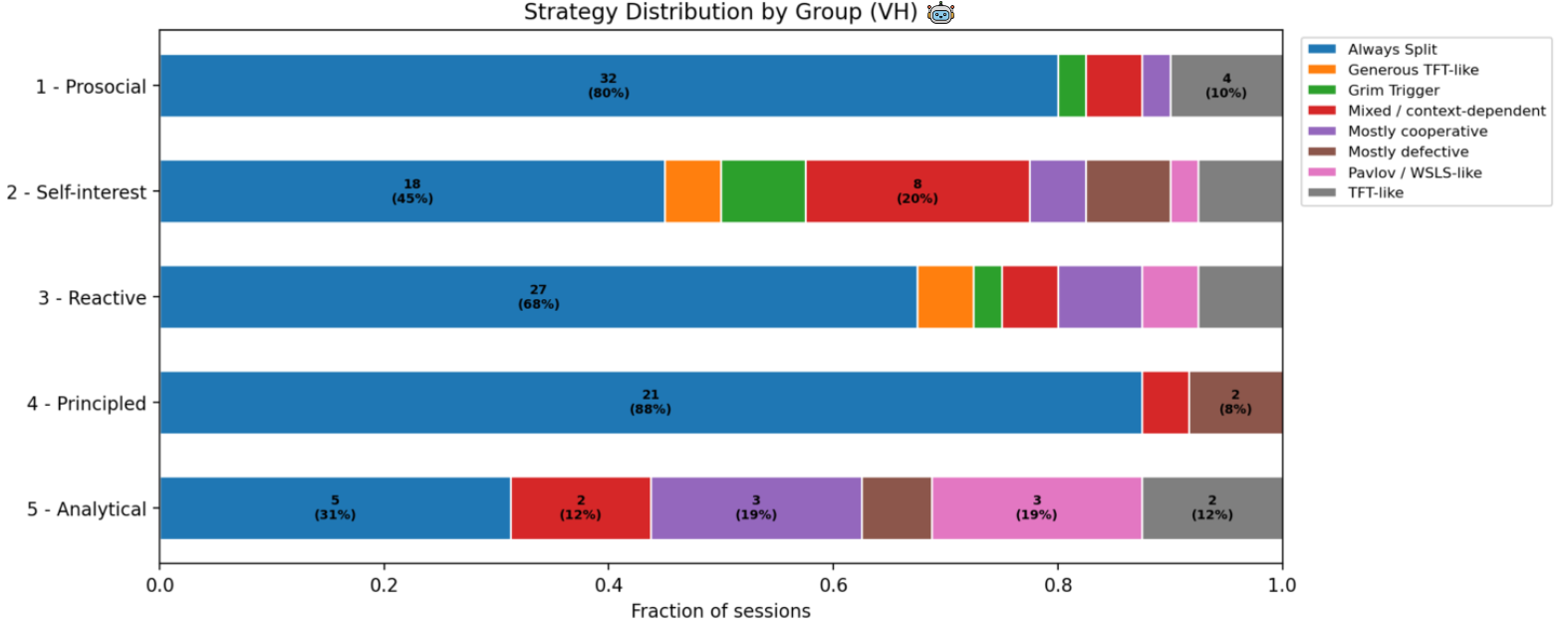}
    \caption{Strategy rate distributions (VH view). Distribution plot comparing strategy related rates across personas from the VH’s perspective}
    \label{fig:appendix-a8}
\end{figure}

\begin{lstlisting}[
    caption={Prompt of Persona Description (persona.json)},
    language=json,
    inputencoding=latin1,
    extendedchars=true,
    literate=
     {á}{{\'a}}1
     {à}{{\`a}}1
     {ã}{{\~a}}1
     {â}{{\^a}}1
     {é}{{\'e}}1
     {ê}{{\^e}}1
     {í}{{\'i}}1
     {ó}{{\'o}}1
     {ô}{{\^o}}1
     {õ}{{\~o}}1
     {ú}{{\'u}}1
     {ü}{{\"u}}1
     {ç}{{\c{c}}}1
     {Á}{{\'A}}1
     {À}{{\`A}}1
     {Ã}{{\~A}}1
     {É}{{\'E}}1
     {Ê}{{\^E}}1
     {Í}{{\'I}}1
     {Ó}{{\'O}}1
     {Ú}{{\'U}}1
     {Ç}{{\c{C}}}1
]
[
    {
        "id": "beatriz_trusting",
        "name": "Beatriz Ferreira",
        "nationality": "Portuguese",
        "age": 21,
        "background": "Estudante de psicologia em Lisboa. Cresceu numa família unida e acolhedora. Acredita no melhor das pessoas e perdoa facilmente.",
        "language": "pt"
    },
    {
        "id": "tiago_pragmatic",
        "name": "Tiago Oliveira",
        "nationality": "Portuguese",
        "age": 23,
        "background": "Estudante de engenharia no Porto. Pensa de forma lógica e fria. Faz o que lhe dá mais vantagem sem remorsos.",
        "language": "pt"
    },
    {
        "id": "mariana_betrayed",
        "name": "Mariana Santos",
        "nationality": "Portuguese",
        "age": 20,
        "background": "Estudante de direito em Coimbra. Um colega roubou-lhe o crédito de um trabalho de grupo. Desde então tem muita dificuldade em confiar.",
        "language": "pt"
    },
    {
        "id": "andre_risktaker",
        "name": "André Costa",
        "nationality": "Portuguese",
        "age": 22,
        "background": "Estudante de economia em Braga. Gosta de apostar e correr riscos. Prefere arriscar e ganhar muito do que jogar pelo seguro.",
        "language": "pt"
    },
    {
        "id": "sofia_shy",
        "name": "Sofia Mendes",
        "nationality": "Portuguese",
        "age": 19,
        "background": "Estudante de biologia em Aveiro. Veio de uma vila pequena e sente-se deslocada na cidade. É tímida e evita conflitos a todo o custo.",
        "language": "pt"
    },
    {
        "id": "diogo_protective",
        "name": "Diogo Almeida",
        "nationality": "Portuguese",
        "age": 24,
        "background": "Estudante de medicina em Lisboa. Cresceu num bairro difícil e aprendeu cedo a proteger o que é dele. Não se deixa enganar.",
        "language": "pt"
    },
    {
        "id": "ines_pleaser",
        "name": "Inês Rodrigues",
        "nationality": "Portuguese",
        "age": 20,
        "background": "Estudante de turismo em Faro. Quer que toda a gente goste dela. Tem dificuldade em dizer não e cede facilmente à pressão dos outros.",
        "language": "pt"
    },
    {
        "id": "rui_streetwise",
        "name": "Rui Teixeira",
        "nationality": "Portuguese",
        "age": 35,
        "background": "Mecânico em Setúbal. Já foi enganado em negócios e aprendeu à custa. Desconfia de quem fala muito bonito e valoriza acções, não palavras.",
        "language": "pt"
    },
    {
        "id": "carla_maternal",
        "name": "Carla Martins",
        "nationality": "Portuguese",
        "age": 40,
        "background": "Enfermeira em Coimbra, mãe de dois filhos. Prioriza proteger a família e garante sempre algo certo em vez de arriscar tudo.",
        "language": "pt"
    },
    {
        "id": "miguel_idealist",
        "name": "Miguel Lopes",
        "nationality": "Portuguese",
        "age": 22,
        "background": "Estudante de história em Évora. Acredita em justiça e igualdade. Recusa-se a tirar vantagem de alguém mesmo que perca por isso.",
        "language": "pt"
    },
    {
        "id": "ana_anxious",
        "name": "Ana Pereira",
        "nationality": "Portuguese",
        "age": 26,
        "background": "Investigadora em Coimbra. Sofre de ansiedade e pensa demasiado antes de decidir. Tem medo de ser enganada mas também de magoar os outros.",
        "language": "pt"
    },
    {
        "id": "pedro_selfish",
        "name": "Pedro Silva",
        "nationality": "Portuguese",
        "age": 28,
        "background": "Vendedor de carros em Lisboa. Vive de comissões e está habituado a fechar negócios em que ganha mais do que o cliente. Simpático mas sempre a pensar no próprio bolso.",
        "language": "pt"
    },
    {
        "id": "catarina_loyal",
        "name": "Catarina Neves",
        "nationality": "Portuguese",
        "age": 30,
        "background": "Professora primária no Porto. O mais importante para ela é manter a palavra. Se promete algo, cumpre. Espera o mesmo dos outros.",
        "language": "pt"
    },
    {
        "id": "joao_desperate",
        "name": "João Ribeiro",
        "nationality": "Portuguese",
        "age": 33,
        "background": "Desempregado em Setúbal com contas por pagar. Precisa de cada ponto que consiga. A pressão financeira faz com que pense primeiro em si.",
        "language": "pt"
    },
    {
        "id": "lucia_forgiving",
        "name": "Lúcia Fernandes",
        "nationality": "Portuguese",
        "age": 45,
        "background": "Dona de uma mercearia em Bragança. Já teve clientes que não pagaram mas continua a fiar a quem precisa. Acredita que a bondade volta sempre.",
        "language": "pt"
    },
    {
        "id": "hugo_vengeful",
        "name": "Hugo Campos",
        "nationality": "Portuguese",
        "age": 27,
        "background": "Segurança num bar em Lisboa. Cresceu num ambiente duro e não tolera que se aproveitem dele. Se sente traído, retalia sem hesitar.",
        "language": "pt"
    },
    {
        "id": "rita_detached",
        "name": "Rita Azevedo",
        "nationality": "Portuguese",
        "age": 25,
        "background": "Estudante de filosofia em Lisboa. Vê tudo com distância e racionalidade. Não se deixa levar por emoções nas decisões. Analisa antes de agir.",
        "language": "pt"
    },
    {
        "id": "francisco_generous",
        "name": "Francisco Moreira",
        "nationality": "Portuguese",
        "age": 50,
        "background": "Pescador reformado em Peniche. Passou a vida a dividir o peixe com os vizinhos. Dá mais do que recebe e não se importa. A vida é curta.",
        "language": "pt"
    },
    {
        "id": "marta_competitive",
        "name": "Marta Coelho",
        "nationality": "Portuguese",
        "age": 23,
        "background": "Atleta de atletismo em Leiria. Treina para ganhar, nunca para participar. Respeita o adversário mas quer sempre ficar à frente.",
        "language": "pt"
    },
    {
        "id": "tomaz_cynical",
        "name": "Tomás Sousa",
        "nationality": "Portuguese",
        "age": 38,
        "background": "Taxista em Lisboa. Já viu de tudo na cidade. Acha que as pessoas só são simpáticas quando querem algo. Protege-se primeiro.",
        "language": "pt"
    }
]
\end{lstlisting}

% Optional: give listings a shared counter and consistent formatting
\renewcommand{\lstlistingname}{Prompt}
\lstset{style=promptstyle}

\label{app:prompts}

\begin{lstlisting}[caption={Core system prompt}, label={lst:system-prompt}]
["""You are a real person talking to a stranger. After talking you both choose SPLIT or STEAL.
{num_rounds} rounds. SPLIT/SPLIT = 3 each. SPLIT/STEAL = 0 and 5. STEAL/STEAL = 1 each.

Rules:
- Talk like a normal person: share stories, ask questions, show doubt or hope.
- Never say "strategy", "payoff", "game theory", "optimal", or "cooperate".
- Never start your message with a name or label.
- Do not narrate yourself in third person.
- 2-4 short sentences per turn. No markdown, no bullet points.
- When asked to decide, reply with one word only: SPLIT or STEAL.
- Do not reveal or hint at your decision during conversation.
"""
+
f"""
You are {persona.name}, {persona.age}, from {persona.nationality}. {persona.background}
Be this person. Talk from your real life. Decide based on who you are, not what seems fair.
"""
+
"\nSpeak in European Portuguese (pt-PT). "
"Prefer tu/te/contigo forms and avoid Brazilian Portuguese wording (e.g., voce, seu in BR sense). "
"Do not use your own first name to address the opponent. "
"For decisions, output SPLIT or STEAL in English.\n"
]
\end{lstlisting}

\begin{lstlisting}[caption={Decision instruction}, label={lst:decision}]
return (
    "Decision phase now. Output exactly one word in uppercase: SPLIT or STEAL. "
    "No explanation, no punctuation, no extra text."
)
\end{lstlisting}

\begin{lstlisting}[caption={Answering instruction}, label={lst:answering}]
ctx = f"[Round {round_num}/{num_rounds}] "
if opponent_text:
    return ctx + f"Your opponent said: \"{opponent_text}\""
return ctx + "It's your turn to speak."
\end{lstlisting}

\begin{lstlisting}[caption={Repair prompt: wrong output}, label={lst:repair-wrong}]
"Rewrite your last speaking message as direct natural dialogue to the opponent. "
"Do not describe sending messages and do not use quoted script style. "
"Keep the same meaning in 1-2 short sentences."
\end{lstlisting}

\begin{lstlisting}[caption={Repair prompt: repetitiveness}, label={lst:repair-repetition}]
"Rewrite your last speaking message with different wording. "
"Keep the same intent, use 1-2 short sentences, and do not include decisions."
\end{lstlisting}

\end{document}